%% file: main.tex
\documentclass[letterpaper,10pt,conference]{ieeeconf}
\IEEEoverridecommandlockouts
\usepackage{amsmath,amssymb,array,booktabs,graphicx,microtype,multirow}
\usepackage{algorithm,algpseudocode}
\usepackage[nocompress]{cite}
\usepackage[hidelinks]{hyperref}
\newcommand{\bibdoi}[1]{doi:~\href{https://doi.org/#1}{\nolinkurl{#1}}}
\usepackage{tikz}
\usetikzlibrary{positioning,arrows.meta,calc,fit}
\title{\LARGE\bf Recovering from Pose-Selection Failures in Repeated Visual Localization}
\author{Jaehyun Kim, Wonseok Kang, and Tae-wan Kim%
\thanks{The authors are with the Department of Naval Architecture and Ocean Engineering, Seoul National University, Seoul, South Korea (e-mail: jaedalong@snu.ac.kr; eoid361@snu.ac.kr; taewan@snu.ac.kr). Corresponding author: Tae-wan Kim.}}

\hypersetup{
  pdftitle={Recovering from Pose-Selection Failures in Repeated Visual Localization},
  pdfauthor={Jaehyun Kim, Wonseok Kang, Tae-wan Kim},
  pdfsubject={HPR preprint}
}

\begin{document}
\flushbottom
\setlength{\parskip}{0ex plus 0.4ex minus 0.1ex}
\bstctlcite{ralIEEEstyle}
\maketitle
\thispagestyle{plain}
\pagestyle{plain}

\begin{abstract}
Repetitive structures can give an incorrect pose candidate more inlier support
than an accurate one, causing visual localization to select it.
History-Guided Pose Reselection (HPR) uses coarse position forecasts from up to five
past localization outputs to distinguish well-separated current-query candidates.
The best-fitting forecast proposes an alternative; HPR accepts it only if every
well-fitting forecast is much closer to it than to the localizer's pose.
HPR needs no odometry and returns the selected candidate unchanged, using
prediction for selection rather than pose refinement.
With one development-set threshold, HPR improves recall within
5 meters and 10 degrees in all 12 main configurations of Hierarchical Localization
(HLoc), MASt3R-based pose generation, and JIST-based retrieval, losing
successes on fewer than 0.4 percent of queries. On NAVER Gangnam B1,
HPR corrects 97 HLoc failures and loses six successes; gains persist within
0.25 meters and 2 degrees on all three HLoc test sets. Median HPR runtime on HLoc
candidates is below one millisecond, excluding pose generation. Recovery weakens under
less frequent localization.
\end{abstract}

\input{sections/final_introduction}
\input{sections/final_related_work}
\input{sections/final_method}
\input{sections/final_setup}

\input{sections/final_results}
\input{sections/final_conclusion}

\bibliographystyle{IEEEtran}
\bibliography{refs}
\end{document}

%% file: sections/final_introduction.tex
\section{Introduction}\label{sec:introduction}
\begin{figure*}[t]
 \centering
 \includegraphics[width=\textwidth]{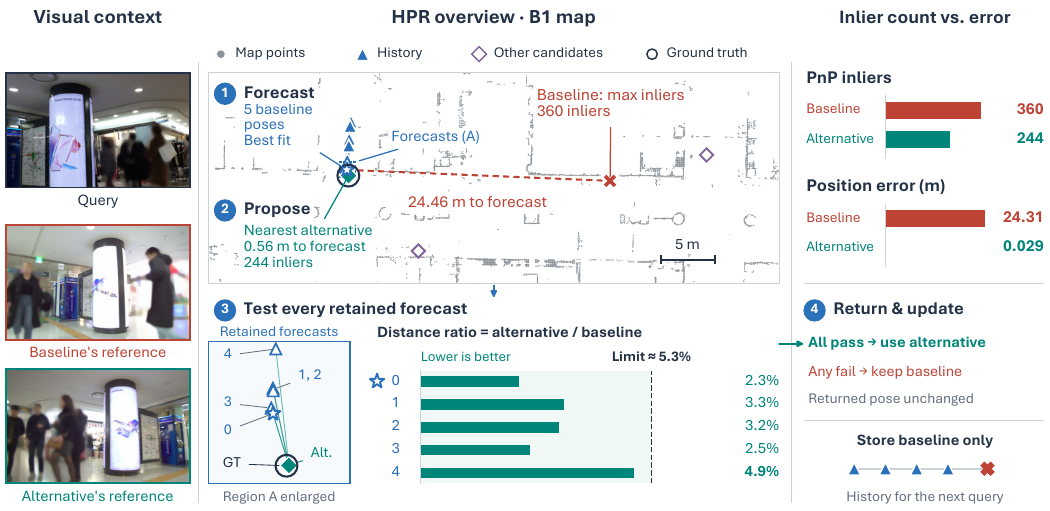}
 \caption{HPR overview on an HLoc B1 query.
 (1--2) The best-fitting forecast ($\star$) proposes the nearest alternative.
 (3) Retained forecasts meet the 5.3\% distance-ratio limit ($\tau\approx0.899$).
 (4) HPR returns the alternative unchanged; history stores the baseline.
 Filled/open triangles mark baseline history/forecasts; labels 0--4 identify
 omitted observations. Region A is enlarged; distances are 3-D.
 Thumbnails show each group's top-retrieved reference. Images, LiDAR, and
 ground truth (GT) are visual context, not HPR inputs.}
 \label{fig:overview}
\end{figure*}

A visual localizer can generate an accurate pose candidate and still select the wrong one:
repetitive structures can give an incorrect candidate more inlier support than
the correct one~\cite{torii2013repetitive,taira2019verification}
(Fig.~\ref{fig:overview}).
With Hierarchical Localization (HLoc)~\cite{sarlin2019coarse}, we estimate a
Perspective-n-Point (PnP) pose per covisibility group and take the
maximum-inlier pose as baseline.
On NAVER Gangnam B1~\cite{lee2021naverlabs}, a correct candidate is already
available for about half of HLoc's failures at 5~m/$10^\circ$.
Recovering these failures is a selection problem.

Pose verification and learned confidence rank candidates by current-query
evidence~\cite{taira2018inloc,taira2019verification,ferranti2021confidence},
which repetition can mislead.
Sequence-based place recognition uses temporal context for reference
matching~\cite{milford2012seqslam,hansen2014hmmvpr,garg2021seqnet,berton2024jist},
while odometry-based verification propagates past estimates~\cite{claxton2024verification}.
We instead use recent localization outputs, without odometry.

We propose History-Guided Pose Reselection (HPR) for repeated localization
along a trajectory, as in robotics and augmented reality.
When candidates are well separated, a coarse position forecast can distinguish
them even if less accurate than the candidate poses.
Because stored outputs can be wrong, HPR fits leave-one-out forecasts to recent
baseline outputs. The best-fitting forecast proposes the nearest alternative, which
replaces the maximum-inlier baseline only if it is much closer under every
well-fitting forecast.
HPR returns the selected candidate unchanged;
accuracy comes from the current query rather than the forecast.

Our contributions are:
\begin{itemize}
  \item HPR, an odometry-free selection module that uses coarse forecasts from recent baseline outputs and returns an existing current-query pose unchanged.
  \item A multi-forecast acceptance test that requires the same alternative to stay preferable to the baseline under every well-fitting forecast. Under the same calibration error budget, HPR exceeds best-fit-only selection on all three HLoc test sets in net recovery at 5~m/$10^\circ$ and 0.25~m/$2^\circ$.
  \item Net 5~m/$10^\circ$ gains in all 12 main configurations with one development-set threshold and few introduced errors; on HLoc, gains also at 0.25~m/$2^\circ$ and after confidence-based selection with HPR recalibrated.
\end{itemize}

%% file: sections/final_related_work.tex
\section{Related Work}\label{sec:related}
HLoc~\cite{sarlin2019coarse} groups references by covisibility, using modular
retrieval~\cite{arandjelovic2016netvlad,keetha2024anyloc} and local features and matching~\cite{detone2018superpoint,sarlin2020superglue,lindenberger2023lightglue}.
Pose verification uses appearance, geometry, or
semantics~\cite{taira2018inloc,taira2019verification}. Ferranti et~al.\ rerank
poses using learned confidence from inlier counts and coverage~\cite{ferranti2021confidence};
Panek et~al.\ select structure-based or structure-less estimates using correspondence
scores~\cite{panek2026combining}. HPR adds past localization outputs as evidence.

Sequence-based visual place recognition (VPR) uses temporal context. SeqSLAM aggregates
sequence evidence~\cite{milford2012seqslam};
SeqNet reranks sequences using learned descriptors~\cite{garg2021seqnet};
Joint Image and Sequence Training (JIST) learns sequence aggregation~\cite{berton2024jist}. Hansen and Browning align sequences
using a hidden Markov model (HMM) and
Viterbi decoding~\cite{hansen2014hmmvpr}. HPR instead uses history to select estimated poses.

Probabilistic filters maintain location beliefs~\cite{doan2021filtering},
some with an off-map state~\cite{xu2021newplace};
Hendrikx et~al.\ track alternative local-to-global feature
associations~\cite{hendrikx2022localglobal}.
Claxton et~al.\ verify past VPR matches and propagate a selected estimate
with odometry~\cite{claxton2024verification}.
HPR stores at most five baseline positions and selects an existing current-query
pose without odometry.

VPR methods also address reference selection, reliability, and calibration.
Bayesian Selective Fusion curates reference sets~\cite{molloy2021bsf}.
Receptiveness prediction assesses sequence matching and considers lower-ranked
alternatives to rejected matches~\cite{hussaini2025smr}, whereas HPR selects
poses using position forecasts.
Malone et~al.\ learn sequence lengths for a desired recall from calibration
data and coarse position priors~\cite{malone2024modulating}.
Spatial uncertainty~\cite{zaffar2024uncertainty} and score-distribution
methods~\cite{miller2026doubt} assess retrieval reliability.
Quantile Transfer carries precision-constrained thresholds from calibration
traversals to deployment by quantile normalization~\cite{rajani2026quantile}.
We test whether HPR's threshold transfers unchanged across sites and pipelines.

%% file: sections/final_method.tex
\section{History-Guided Pose Reselection}\label{sec:method}
\subsection{Problem Formulation and Overview}
HPR uses coarse forecasts to select an existing current-query pose. It proposes
the alternative nearest the best-fitting forecast, compares that alternative
with the baseline under every well-fitting forecast, and returns the selected pose unchanged
(Fig.~\ref{fig:overview}, steps 1--4).

Let $\mathcal B_t=\{T_t^0,\ldots,T_t^{K_t-1}\}$ contain $K_t$
current-query poses, with the maximum-inlier baseline indexed by~0.
Their camera centers $c_t^j\in\mathbb R^3$ share a common metric map frame.
Missing estimates and numerical failures (e.g., non-finite centers) are unavailable;
missing baselines fail.
Write $b_t=c_t^0$ and let $\mathcal J_t$ index available alternatives.
The output is $\widehat T_t\in\{T_t^0\}\cup\{T_t^j:j\in\mathcal J_t\}$.

Before query $t$, each camera/view stream stores its latest $m\le5$ baseline
centers in $\mathcal H_t=(h_0,\ldots,h_{m-1})$, oldest first.

\subsection{Robust Forecasting from Baseline-Pose History}
Under approximately constant motion and regular output intervals, leave-one-out fits reduce
sensitivity to an inconsistent history position.
With five stored centers indexed by $i=0,\ldots,4$, each fit omits $h_k$ and fits
$h_i\approx\beta_0+i\beta_1$ by least squares to the other four at their original
indices, then predicts index~5:
\begin{equation}
 p_k=\beta_0+5\beta_1, \qquad \beta_0,\beta_1\in\mathbb R^3.
 \label{eq:loo}
\end{equation}
Indices count stored observations, not elapsed time; forecasts can become stale after gaps.
Let $r_k$ be the root-mean-square Euclidean fit residual on the four included
positions, in meters. The fit with smallest $r_k$, indexed by $k_*$,
defines the best-fitting forecast $p_*=p_{k_*}$.
This is least-trimmed-squares (LTS) fitting with squared Euclidean
residuals~\cite{rousseeuw1987robust}.

HPR retains well-fitting forecasts according to their residuals relative to the best fit:
\begin{equation}
 \mathcal S_t=\{p_*\}\cup
 \{p_k:r_k\le 2r_{k_*}+0.05\,\mathrm m+\varepsilon\}.
 \label{eq:scenarios}
\end{equation}
Testing the retained forecasts checks whether the same alternative remains
preferable to the baseline across similarly fitting history subsets.
The 0.05~m slack retains near-equivalent fits when the best residual is near zero.
Here $\varepsilon=10^{-9}$ is a numerical tolerance.
These settings are fixed across datasets.

With three or four stored centers, the fallback extends the last three centers'
average displacement by one observation:
\begin{equation}
 p_*=h_{m-1}+\tfrac12(h_{m-1}-h_{m-3}),
 \qquad \mathcal S_t=\{p_*\}.
 \label{eq:fallback}
\end{equation}
HPR keeps the baseline if fewer than three positions are available (warm-up) or no
forecast can be formed.

\subsection{Candidate Proposal and Replacement Testing}
The forecast $p_*$ proposes the nearest alternative:
\begin{equation}
 j_* =\arg\min_{j\in\mathcal J_t}
       \|c_t^j-p_*\|,
 \qquad a_t=c_t^{j_*}.
 \label{eq:nominee}
\end{equation}
For a retained forecast $p$, the margin is positive when the alternative is closer:
\begin{equation}
 g(b,a;p)=\frac{\|b-p\|-\|a-p\|}
                   {\|b-p\|+\|a-p\|+\varepsilon}.
 \label{eq:margin}
\end{equation}
HPR requires a margin of at least
$\tau$ under every retained forecast:
\begin{align}
 G_t&=\min_{p\in\mathcal S_t}g(b_t,a_t;p),\label{eq:gate}\\
 \widehat T_t&=
 \begin{cases}
 T_t^{j_*},&\text{if eligible and }G_t\ge\tau,\\
 T_t^0,&\text{otherwise}.
 \end{cases}\label{eq:output}
\end{align}
A query is eligible when the baseline, an alternative, and a retained forecast
are available. The candidate centers must differ ($\|b_t-a_t\|>\varepsilon$),
and the sum of their distances to $p_*$ must exceed the tolerance
($\|b_t-p_*\|+\|a_t-p_*\|>\varepsilon$).

Acceptance requires the alternative to be at most $(1-\tau)/(1+\tau)$ times as far
from every retained forecast as the baseline (ignoring $\varepsilon$).
The dimensionless threshold $\tau\approx0.899$, selected on the development set,
gives a ratio of about 5.3\%
(Fig.~\ref{fig:overview}), not a probability of correctness.

Because $p_*$ is retained, $G_t\le g(b_t,a_t;p_*)$: at any threshold, this
multi-forecast test accepts a subset of the replacements that $p_*$ alone would accept,
and thus introduces no more errors. The same error budget can therefore admit a lower
$\tau$ than testing $p_*$ alone.

\subsection{History Update and Calibration}\label{sec:cma_update}
After each decision, HPR stores the latest five available baseline centers;
history starts empty and persists across missing outputs.
Because history stores only baseline centers, a wrong replacement cannot
reinforce itself through later forecasts.
For fixed candidate sets, the forecasts, proposed alternatives, and margins
therefore do not depend on $\tau$.
One pass computes calibration margins; raising $\tau$ only removes replacements.

%% file: sections/final_setup.tex
\section{Experimental Setup}\label{sec:setup}
We test recovery across 12 configurations with one HLoc-B2 threshold.
Component tests isolate the mechanism; selector and confidence comparisons
assess performance and complementarity. Calibration and history analyses
characterize operating limits.

\subsection{Datasets and Sequential Evaluation}
We use NAVER Gangnam B1/B2 and Hyundai B1~\cite{lee2021naverlabs}, nighttime
360Loc (night)~\cite{huang2024360loc}, and OpenLORIS Corridor~\cite{shi2020openloris}
(Table~\ref{tab:streams}).
Two Gangnam B2 captures form the development set for threshold calibration.
The HLoc test sets (targets) assess transfer between floors (B1) and sites
(night and Hyundai). NAVER uses public-validation ground truth (GT).
We replay queries chronologically and score every query, including missing
poses and warm-up. Each capture and camera/view keeps its own history,
which persists across gaps.

We additionally apply the unchanged HLoc-B2 threshold to LaMAR public validation
(CAB\slash HGE\slash LIN, phone\slash HoloLens)~\cite{sarlin2022lamar}.
Each capture and camera/rig keeps its own history, with one selection and update per rig;
extrinsics convert selected rig poses to camera poses.

\input{hpr_streams_table}

\subsection{Pose Generation}
We test pipeline transfer in 12 configurations: HLoc, MASt3R, and DUNE+MASt3R (DUNE) on B1, night,
and Hyundai, and three JIST retrieval settings on OpenLORIS (Table~\ref{tab:transfer}).
For all candidate streams, the baseline is the pose with the most inliers.
On NAVER and 360Loc, HLoc retrieves 200 reference images with MegaLoc~\cite{berton2025megaloc},
matches them to the query using SuperPoint/LightGlue,
and estimates a PnP pose per covisibility group.
We test retrieval depth with HLoc top-100 retrieval on B1.
On LaMAR, HLoc uses MegaLoc top-10 retrieval and SuperPoint/SuperGlue, with
PnP poses for each covisibility group.

MASt3R and DUNE~\cite{mast3r2024,sariyildiz2025dune}
group MegaLoc references by camera-center distance. Known reference poses place
their predicted metric point maps in map coordinates, without scale fitting.
Dense query-to-map matches feed SQPnP~\cite{terzakis2020sqpnp} with RANSAC per group.

JIST tests whether HPR adds recovery with sequence-based retrieval on OpenLORIS.
Sessions 2--5 query session 1 using the released JIST~\cite{berton2024jist}
(ResNet-18, 512-D) descriptors, raw or pooled with Sequential Generalized Mean
(SeqGeM) over one ($T{=}1$)
or up to five causal frames ($T{\leq}5$), for queries and references.
The last frames of JIST's top-200 retrieved windows form camera-center groups;
SuperPoint/SuperGlue matching and PnP give one pose per group.

\subsection{Metrics and Threshold Calibration}\label{sec:threshold_selection}
We measure recovery from incorrect-place selections at 5~m/$10^\circ$
(primary) and precise localization at 0.25~m/$2^\circ$
(strict)~\cite{sattler2018benchmarking}.
On NAVER and LaMAR, we also rescore unchanged outputs at their official
0.1~m/$1^\circ$ and 1~m/$5^\circ$ criteria~\cite{lee2021naverlabs,sarlin2022lamar}.
Position and rotation bounds are inclusive; missing poses fail.
A baseline failure is recoverable if at least one current-query candidate meets the criterion.
Corrections $C$ turn failures into successes; introduced errors $H$ turn
successes into failures. Tables report $C/H$; net recall change is $(C-H)/N$
over $N$ queries.

Each thresholded selector maximizes B2 corrections under the empirical
error budget $H\leq5$ (0.55\% of B2 queries).
Margin thresholds are inclusive; calibration searches nonnegative margins and a no-replacement option.
Ties prefer fewer introduced errors, then fewer replacements, then a higher threshold.
Unless recalibration is stated, B2 thresholds transfer unchanged, and strict
evaluation rescores the same returned poses without retuning.
Rate-matched limits $H\leq\lfloor5N/916\rfloor$ scale the empirical budget to other
calibration sets of $N$ queries, giving $H\leq2$ on one B2 capture.
These budgets do not guarantee target error rates.

\subsection{Compared Selectors}
\input{hpr_expanded_transfer_table}
Comparisons share each configuration's candidates (Table~\ref{tab:primary}).
\emph{Current-query selectors} test selection without history.
Thresholded variants use $(s_a-s_b)/(s_a+s_b+\varepsilon)$,
where $s_a,s_b$ score the highest-scoring alternative and baseline.
Inlier ratio uses $s_j=n_j/m_j$: PnP inliers
$n_j$ over input 2\nobreakdash-D--3\nobreakdash-D correspondences $m_j$.
We adapt Ferranti et~al.'s logistic confidence~\cite{ferranti2021confidence}
using inlier count and query/reference image coverage, taking maximum per-image
reference coverage for groups. We compare maximum confidence with calibrated
score-difference selection.
We fit confidence and its selection threshold on HLoc B2
at 5~m/$10^\circ$.
Confidence comparisons use HLoc, JIST, and MASt3R/DUNE
Hyundai candidates.

\emph{Temporal controls} test simpler forecasts and other uses of localization history.
Constant-position and constant-velocity controls~\cite{barshalom2001estimation}
predict $p=h_{m-1}$ and $p=2h_{m-1}-h_{m-2}$ from $m$ stored
positions; both test the nearest alternative with a single-forecast margin.
Our full-history HMM uses inlier support and Gaussian transition weights
without row normalization:
\begin{equation*}
\pi_t(j)\propto(1+n_{t,j})\sum_i\pi_{t-1}(i)
\exp\!\left(-\frac{\|c_t^j-c_{t-1}^i\|^2}{2\sigma^2}\right),
\end{equation*}
where $\pi_t(j)$ is the belief assigned to candidate $j$, $n_{t,j}$ is its
inlier count, and $\sigma$ is the displacement scale.
Beliefs are normalized before the same score-difference test;
history resets when no center is available.
B2 selects $\sigma=8$\,m by the area under its correction--error curve.
We evaluate both unnormalized and row-normalized transitions. Each either fixes
$\sigma=8$\,m or jointly tunes $\sigma\in\{0.25,0.5,1,2,4,8,16\}$\,m and the
threshold under the B2 error limit; both joint tunings select $\sigma=4$\,m.

Kalman controls use HPR's history (at least three positions), unit-step constant
velocity, measurement covariance $R=I_3$, and acceleration-noise scale $\eta$.
Predicted position $p^-$ has covariance $P^-_{pp}$; normalized innovation
squared (NIS) $\nu_j^2$ uses $P^-_{pp}+R$~\cite{barshalom2001estimation}.
Kalman/NIS accepts the minimum-NIS alternative at threshold $\gamma$ when $\nu_a^2\le\gamma<\nu_b^2$;
Kalman/margin tests the nearest alternative with HPR's margin. B2 tunes
thresholds and $\eta$ independently; both select $\eta=1$.

\emph{HPR ablations} keep HPR's history, fallback, and eligibility but change
fitting or acceptance. Least squares fits all five positions and both proposes
and tests with that single forecast. LTS-only, the best-fit-only version, tests
only $p_*$. Separately calibrated thresholds compare methods under the same
B2 budget. Sharing HPR's threshold isolates the multi-forecast test;
both propose the same alternative.
Unless stated otherwise, LTS-only uses its separately calibrated threshold.
\input{hpr_combined_primary_strict_table}

%% file: hpr_streams_table.tex
\begin{table}[t]
\centering
\small
\caption{Query Streams and Five-Position History Spans}
\label{tab:streams}
\begin{tabular*}{\columnwidth}{@{\extracolsep{\fill}}lrrr@{}}
\toprule
Dataset & Queries & Streams & Median span\\
\midrule
Gangnam B2~\cite{lee2021naverlabs} & 916 & 6 & 8~s\\
Gangnam B1~\cite{lee2021naverlabs} & 2,620 & 8 & 8~s\\
360Loc night~\cite{huang2024360loc} & 6,982 & 12 & 4 frames\\
Hyundai~\cite{lee2021naverlabs} & 1,751 & 4 & 8~s\\
OpenLORIS Corridor~\cite{shi2020openloris} & 786 & 4 & 2~s\\
LaMAR~\cite{sarlin2022lamar} & 2,732 & 19 & 4~s\\
\bottomrule
\end{tabular*}
\par\smallskip\parbox{\columnwidth}{\footnotesize 360Loc uses two views per sequence.
LaMAR counts frame/rig updates, evaluated on 5,263 camera images.}
\end{table}

%% file: hpr_expanded_transfer_table.tex
\begin{table}[!t]
\centering\small
\setlength{\tabcolsep}{2pt}\renewcommand{\arraystretch}{1.0}
\caption{Pose Recovery Across Configurations}
\label{tab:transfer}
\begin{tabular*}{\columnwidth}{@{\extracolsep{\fill}}l*{5}{c}@{}}
\toprule
& \multicolumn{2}{c}{Baseline successes} & Recoverable
& \multicolumn{2}{c}{\textbf{HPR} $C/H$}\\
\cmidrule(lr){2-3}\cmidrule(lr){5-6}
Configuration & Primary & Strict & $A/F$ & Primary & Strict\\
\midrule
HLoc B1 & 2,200 & 1,886 & 210/420 & 97/6 & 66/5\\
HLoc night & 6,037 & 3,747 & 121/945 & 76/1 & 18/0\\
HLoc Hyundai & 1,714 & 1,469 & 21/37 & 16/0 & 14/0\\
\midrule
MASt3R B1 & 2,400 & 269 & 69/220 & 38/3 & 4/0\\
MASt3R night & 3,813 & 65 & 440/3,169 & 16/3 & 0/0\\
MASt3R Hyundai & 1,686 & 194 & 46/65 & 8/0 & 1/0\\
\midrule
DUNE B1 & 2,419 & 1,034 & 48/201 & 23/2 & 6/1\\
DUNE night & 4,181 & 33 & 556/2,801 & 14/0 & 0/0\\
DUNE Hyundai & 1,707 & 237 & 22/44 & 3/0 & 0/0\\
\midrule
JIST raw & 549 & 484 & 43/237 & 11/3 & 7/3\\
JIST $T{=}1$ & 549 & 483 & 33/237 & 10/2 & 9/2\\
JIST $T{\leq}5$ & 541 & 474 & 45/245 & 11/3 & 6/3\\
\bottomrule\end{tabular*}
\par\smallskip\parbox{\columnwidth}{\footnotesize HLoc-B2 $\tau$; baseline-only history.
$A/F$: recoverable/all primary baseline failures; $C/H$: corrections/introduced errors.}
\end{table}

%% file: hpr_combined_primary_strict_table.tex
\begin{table*}[t]
\centering\footnotesize
\setlength{\tabcolsep}{2pt}
\caption{HLoc Pose Selection at Primary and Strict Criteria}
\label{tab:primary}\label{tab:strict}
\begin{tabular*}{\textwidth}{@{\extracolsep{\fill}}lcc*{3}{cc}@{}}
\toprule
 & & B2 (dev.) & \multicolumn{2}{c}{B1} & \multicolumn{2}{c}{360Loc night} & \multicolumn{2}{c}{Hyundai}\\
\cmidrule(lr){4-5}\cmidrule(lr){6-7}\cmidrule(lr){8-9}
Method & $\tau$ & Primary & Primary & Strict & Primary & Strict & Primary & Strict\\
\midrule
Baseline & -- & $66.92\ (0/0)$ & $83.97\ (0/0)$ & $0/0$ & $86.47\ (0/0)$ & $0/0$ & $97.89\ (0/0)$ & $0/0$\\
Confidence (max.) & -- & $77.40\ (109/13)$ & $\mathbf{88.55}\ (131/11)$ & $99/6$ & $87.01\ (67/29)$ & $37/21$ & $97.49\ (15/22)$ & $12/22$\\
Confidence (cal.) & 0.1512 & $76.31\ (91/5)$ & $88.24\ (115/3)$ & $90/1$ & $86.94\ (48/15)$ & $22/7$ & $97.60\ (14/19)$ & $11/20$\\
Inlier ratio & 0.8376 & $66.59\ (2/5)$ & $82.52\ (2/40)$ & $1/27$ & $85.49\ (0/68)$ & $1/24$ & $97.09\ (2/16)$ & $2/6$\\
\specialrule{.3pt}{1pt}{1pt}
Constant position & 0.9846 & $67.90\ (14/5)$ & $84.50\ (18/4)$ & $9/4$ & $86.88\ (29/0)$ & $8/0$ & $98.00\ (3/1)$ & $2/0$\\
Constant velocity & 0.9924 & $67.79\ (13/5)$ & $84.77\ (21/0)$ & $14/0$ & $86.85\ (27/0)$ & $9/0$ & $98.00\ (2/0)$ & $1/0$\\
Full-history HMM (8 m) & 0.6581 & $78.49\ (111/5)$ & $88.40\ (154/38)$ & $109/25$ & $86.52\ (80/76)$ & $16/35$ & $98.69\ (14/0)$ & $11/0$\\
Full-history HMM (joint) & 0.9989 & $79.48\ (120/5)$ & $86.56\ (153/85)$ & $108/65$ & $85.75\ (79/129)$ & $15/54$ & $98.74\ (15/0)$ & $11/0$\\
HMM (normalized, 8 m) & 0.9846 & $75.76\ (85/4)$ & $\mathbf{88.55}\ (131/11)$ & $91/7$ & $86.82\ (60/35)$ & $14/17$ & $98.69\ (14/0)$ & $11/0$\\
HMM (normalized, joint) & 0.9987 & $78.38\ (110/5)$ & $87.02\ (150/70)$ & $104/53$ & $86.05\ (78/107)$ & $16/55$ & $98.74\ (15/0)$ & $11/0$\\
Kalman/NIS & -- & $69.76\ (31/5)$ & $86.30\ (66/5)$ & $48/3$ & $87.11\ (50/5)$ & $16/4$ & $98.23\ (8/2)$ & $6/2$\\
Kalman/margin & 0.9451 & $69.43\ (26/3)$ & $86.18\ (60/2)$ & $42/2$ & $87.07\ (42/0)$ & $12/0$ & $98.40\ (9/0)$ & $7/0$\\
\specialrule{.3pt}{1pt}{1pt}
Least squares & 0.9439 & $68.78\ (22/5)$ & $86.03\ (57/3)$ & $41/3$ & $87.07\ (43/1)$ & $11/0$ & $98.46\ (10/0)$ & $8/0$\\
LTS-only (separately calibrated) & 0.9575 & $70.31\ (35/4)$ & $86.56\ (71/3)$ & $50/3$ & $87.44\ (70/2)$ & $16/0$ & $98.34\ (8/0)$ & $6/0$\\
LTS-only (HPR threshold) & 0.8991 & $71.07\ (52/14)$ & $87.40\ (98/8)$ & $66/7$ & $\mathbf{87.54}\ (77/2)$ & $19/0$ & $\mathbf{98.80}\ (16/0)$ & $14/0$\\
\textbf{HPR} & 0.8991 & $71.72\ (49/5)$ & $87.44\ (97/6)$ & $66/5$ & $\mathbf{87.54}\ (76/1)$ & $18/0$ & $\mathbf{98.80}\ (16/0)$ & $14/0$\\
\bottomrule\end{tabular*}
\par\vspace{1pt}
\parbox{\textwidth}{\footnotesize Primary: recall (\%) and $C/H$ at 5~m/$10^\circ$. Strict: $C/H$ at 0.25~m/$2^\circ$ for the same outputs.
Bold: highest recall per test set.
Learned-model B2 counts are in-sample. Confidence max. uses no threshold.
LTS-only rows compare operating points or isolate the multi-forecast test.
Kalman/NIS: $\gamma=0.7208$.}
\end{table*}

%% file: sections/final_results.tex
\begin{figure*}[t]
 \centering
 \includegraphics[width=0.92\textwidth,trim=0 5bp 0 5bp,clip]{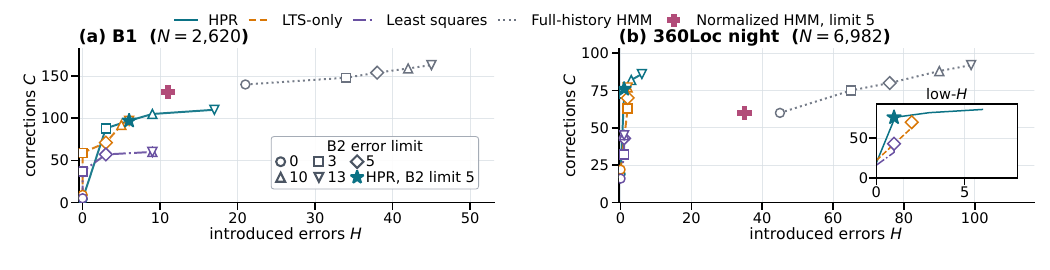}
 \caption{Correction--error trade-offs at 5~m/$10^\circ$.
 Markers show operating points selected under increasing B2 error limits and
 transferred unchanged to targets.
 HMMs use $\sigma=8$\,m; stars and filled crosses mark HPR and row-normalized HMM at limit 5.
 The night inset enlarges low-error points.}
 \label{fig:tradeoff}
\end{figure*}

\section{Results and Discussion}\label{sec:results}
\subsection{Recovery Across Configurations}
With one B2 threshold, HPR improves primary recall in all 12 configurations
of Table~\ref{tab:transfer}, with introduced-error rates of 0--0.38\% of queries.
On HLoc B1, HPR corrects 97 of 210 recoverable failures (46\%), introduces
six errors, and gains 3.47 percentage points of recall
(Tables~\ref{tab:transfer}--\ref{tab:primary}, Fig.~\ref{fig:tradeoff}).
Gains persist in both B1 captures and all six night sequences.

\begin{figure}[t]
 \centering
 \includegraphics[width=\columnwidth]{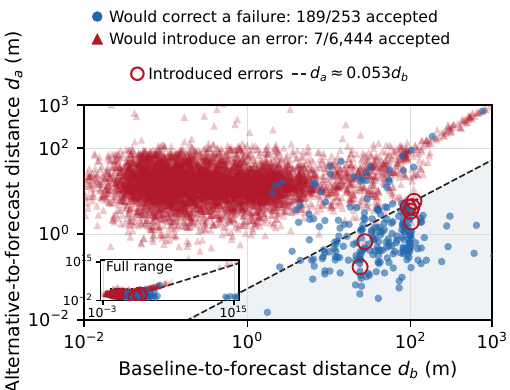}
 \caption{All 6,697 HLoc test-set proposals that would change primary success status.
 $d_a$ and $d_b$ are the alternative's and baseline's distances to the minimum-margin forecast.
 The shaded region below $d_a\approx0.053d_b$ satisfies $G_t\geq\tau$; rings mark
 the seven introduced errors. The inset includes out-of-range proposals, with six
 corrections from extreme baselines (screening: Table~\ref{tab:calibration-robustness}(b)).}
 \label{fig:margin-mechanism}
\end{figure}

HPR accepts replacements selectively: over HLoc targets, it accepts 189 of 253
correcting proposals but only seven of 6,444 error-inducing proposals,
recovering 189 of 352 recoverable failures (54\%; Fig.~\ref{fig:margin-mechanism}).

The same returned poses improve strict recall on all three HLoc targets, with no
introduced errors on night or Hyundai (Table~\ref{tab:strict}).
At NAVER's official 0.1~m/$1^\circ$ and 1~m/$5^\circ$ criteria, the same outputs add 28 and 86 net
successes on HLoc B1, and nine and 16 on Hyundai, respectively.

Recovery extends to dense pipelines and sequence-based retrieval.
Strict gains occur in nine of the 12 configurations;
the remaining three, MASt3R night, DUNE night, and DUNE Hyundai, have only two,
two, and one recoverable strict failures, respectively.
On OpenLORIS, HPR adds recovery to all three JIST settings at both criteria.
With JIST $T{\leq}5$, HPR loses two primary and three strict net successes in session 2.
Top-100 retrieval retains most B1 recovery at the unchanged threshold: $93/10$
(primary) and $55/6$ (strict).

\phantomsection\label{sec:runtime}
HPR's $O(K_t)$ selection with history updates takes 0.44--0.72~ms median
on HLoc (99th percentile: 0.72--0.97~ms) on one AMD Ryzen~7 5800X core,
excluding pose generation.

\subsection{Multiple Forecasts and Other Design Choices}\label{sec:multi-forecast}
Testing multiple forecasts can veto a replacement supported by the best fit
(Table~\ref{tab:primary}, last three rows).
On 898 eligible B2 queries with alternatives fixed, at $\tau=0.8991$ the
multi-forecast test cuts LTS-only's 14 introduced errors to five while losing
three of 52 corrections. LTS-only needs $\tau=0.9575$ to meet $H\leq5$,
leaving 35 corrections.

Under the same B2 error budget, HPR gains 91 net primary B1 successes versus LTS-only's 68.
At HPR's threshold, LTS-only closes 22 of this 23-success gap and all of the night/Hyundai gap,
but exceeds the B2 error budget.
At this shared threshold, the multi-forecast test withholds two B1 errors and one
B1 correction: one net success, with 20 outputs altered across HLoc targets.
Most of this B2-calibrated advantage therefore comes from the lower threshold
the multi-forecast test permits;
the test's direct effect grows at lower night- and Hyundai-calibrated thresholds.

HPR also exceeds LTS-only by 2--14 net primary successes
in all six MASt3R/DUNE configurations and is within one net
success of LTS-only at both criteria on JIST.

Robust fitting helps on two of three targets: LTS-only improves on least squares
on B1 and night, but not Hyundai, at both criteria (Table~\ref{tab:primary}).
Alternative selection matters little: max-min selects the eligible alternative
with the highest minimum margin and, at its own B2-calibrated threshold, gives
B1 $96/6$ versus HPR's $97/6$ and equal night/Hyundai $C/H$.
Forecasting variants change one setting: window length, residual factor or slack
(\ref{eq:scenarios}), or time indexing. They retain net gains at both criteria
(primary counts in Table~\ref{tab:calibration-robustness}(a)); time indexing
uses NAVER timestamps and 360Loc frames.
\phantomsection\label{sec:candidate-screening}\label{sec:screened-results}%
Screening extreme candidates preserves HPR's advantage over LTS-only at both
criteria after B2 recalibration (Table~\ref{tab:calibration-robustness}(b)).
Screening excludes numerical failures, centers beyond 1,000~m from their reference
cameras' centroid, and rotations failing orthonormality or unit determinant checks.
No surviving candidate means a missing output.
\input{hpr_sensitivity_table}

\subsection{Comparison with Other Selectors}
Among selectors calibrated under the B2 error budget, HPR has the highest net
recovery on night and Hyundai at both criteria (Table~\ref{tab:primary}). Both HMMs
with $\sigma=8$\,m exceed HPR's net B1 recovery (normalized HMM: 120 versus 91),
but introduce more errors on night.
Both HMMs also have lower net night recovery and more B1 errors
(Fig.~\ref{fig:tradeoff}); joint tuning increases these errors.
Constant-position/velocity and Kalman controls recover fewer failures than HPR.

Among current-query scores, inlier ratio loses net successes on every HLoc target. HPR outperforms confidence
at both criteria on HLoc night and Hyundai, but not on HLoc B1 or the three JIST settings.
Without refitting, the HLoc-B2-trained confidence model loses most baseline successes
on MASt3R/DUNE Hyundai, where HPR still gains (Tables~\ref{tab:transfer} and~\ref{tab:confidence-transfer}(a)).

\subsection{Complementarity with Confidence Selection}
History complements current-query confidence after recalibration.
We test a cascade: confidence outputs become HPR's baselines and history.
We retain HPR's threshold or recalibrate it on B2, maximizing combined
corrections under five total introduced errors against maximum inliers.
Recalibrated HPR changes confidence's B1 $C/H$
from $115/3$ to $138/3$ and night's from $48/15$ to $69/7$
(Table~\ref{tab:confidence-transfer}(a)); on Hyundai, it removes two errors.
HPR alone has higher primary recall than the recalibrated cascade on night and Hyundai.
Reusing HPR's threshold gives more recovery but six B2 errors against a limit of five.
Both settings add net JIST recovery over confidence alone.

With confidence fitted on one B2 capture and calibrated on the other, HPR still raises
net recovery for every direction, limit, and target (Table~\ref{tab:confidence-transfer}(b)).
On B1, HPR adds errors in the $1\to2$ direction but removes them in $2\to1$.
In $2\to1$, HPR reduces but does not remove confidence's Hyundai losses.

\input{hpr_confidence_transfer_table}

\subsection{Calibration Transfer and Uncertainty}
\label{sec:calibration-robustness}
HPR exceeds separately calibrated LTS-only in 11 of 12 primary
cross-calibration directions.
Both methods are calibrated on each source (B2, B1, night, or Hyundai) at its
rate-matched limit and evaluated on the other three
(Table~\ref{tab:calibration-robustness}(c)).
All four calibration datasets yield lower HPR thresholds than LTS-only's; only
Hyundai-to-night reverses the primary comparison. No strict comparison reverses.
When both methods use HPR's night- or Hyundai-calibrated threshold, the
multi-forecast test adds 38--50 net primary successes over LTS-only on B2 and B1,
but HPR then makes 58 and 61 B1 errors, respectively, exceeding B1's
rate-matched limit of 14.

Calibrated on a single B2 capture under $H\leq5$ or the rate-matched $H\leq2$, HPR leads
LTS-only by 2--25 net primary and 1--14 strict successes per HLoc test set;
$H\leq2$ gives HPR fewer B1/night errors than $H\leq5$.
DUNE-B2 recalibration gives $31/0$ on night versus LTS-only's $7/0$,
without strict gains. On DUNE B1, HPR's $23/2$ becomes $30/6$;
recalibrated LTS-only gives $23/2$.

Fixed-threshold gains survive target resampling: 4,000 paired stationary block-bootstrap
replicates~\cite{politis1994stationary} per block length resample
simul\-ta\-ne\-ous-view groups (views sharing timestamps or source frames) within each
capture or source sequence, with thresholds fixed and weights proportional to the
original query counts. For $M_s$ groups in sequence $s$, mean block
lengths are $\{0.5,1,2,4\}\lceil\sqrt{M_s}\rceil$.
At each method's fixed B2 threshold, all 95\% primary bootstrap intervals favor
HPR across targets (Table~\ref{tab:calibration-robustness}(d)) and block lengths;
only night's strict intervals include zero.

Calibration resampling instead recalibrates the HPR and LTS-only thresholds
under $H\leq5$: 1,000 paired B2 resamples per length preserve groups and each
capture's query share. Histories and targets remain fixed.
For each target and block length, HPR has higher net primary recovery in 84--93\%
of these resamples, but the recalibrated ranges in Table~\ref{tab:calibration-robustness}(d)
also allow higher net recovery for LTS-only on each target.

LaMAR transfer is mixed: HPR gains four net primary successes and none at the strict
criterion (Table~\ref{tab:lamar}); all introduced errors occur on HGE HoloLens,
which loses at all four criteria (primary introduced-error rate: 1.03\%).
\input{hpr_lamar_table}

\subsection{Operating Conditions and Error Analysis}
\label{sec:geometry}
Candidate separation and informative recent history shape HPR's observed recovery.
Corrected queries have median baseline--alternative separations of 40--82~m,
versus 0.25--1.05~m from the proposing forecast to the alternative.
Cadence, outage, past-GT, and offset analyses keep candidates and B2-calibrated thresholds fixed.
Less frequent localization reduces B1 recovery. Subsampling HLoc streams at strides
1, 2, 4, and 8 (each starting offset a separate stream with its own history and warm-up)
lowers net gains from 91 to four as median intervals grow from 2.0 to 16.0~s
(Table~\ref{tab:operating}). At stride 8, two of eight starting offsets lose
successes. Night and Hyundai gains shrink but remain positive.
Across HLoc targets, gaps over 1.5 times the median history interval affect
0.98\% of post-warm-up queries (2.50\% on B1),
without introduced errors.
\input{hpr_operating_table}

Periodic outages retain whole-stream gains, but the first query after an outage is at risk.
With outputs periodically suppressed (1, 3, 5, or 10 every 50 queries, at offsets
10, 25, or 40 per stream), default HPR still gains 62--90, 52--76, and 8--16
net primary successes on B1, night, and Hyundai, with at most nine errors per test set;
missing-output resets, which wait for five fresh positions, remove 0--3 errors
but sacrifice 0--11 corrections. Adding resets after long gaps loses four B1
and three Hyundai corrections on unmodified streams without removing errors.

Resumption tests score each of 2,500 B1 queries separately, after omitting
its 1, 3, 5, or 10 preceding outputs. Every scored query follows an outage.
On these queries, omitting ten outputs changes primary $C/H$ from $94/6$ (no
omission) to $5/17$; timestamp-based fitting gives $9/4$.

Replacing only past positions with ground truth removes all seven HLoc
introduced errors (Table~\ref{tab:operating}).
These errors follow two to five wrong stored positions, and after each, HPR
outputs a correct pose within three queries. Such errors can arise when a shared
offset in past positions shifts every forecast without changing fit residuals.

Injected offsets mainly reduce corrections. We add a shared random 3-D
offset (5, 20, or 50~m; three seeds) to 1, 3, or 5 consecutive stored positions
every 50 baseline updates, starting at update 11.
Even five consecutive offsets, which affect more queries than shorter bursts,
leave net primary gains of 78--81, 61--70, and 14--15 on B1, night, and Hyundai,
with at most seven introduced errors per test set.
Here, HPR exceeds separately calibrated LTS-only but can trail LTS-only at
HPR's threshold by two net primary successes.

\phantomsection\label{sec:history-comparison}
Feedback adds corrections but can lengthen error runs.
Full feedback stores returned poses after selection; its threshold is
recalibrated by sequential B2 replay. It gives $156/2$ on B1 but $15/22$ on
JIST $T{\leq}5$, including 13 consecutive introduced errors.
Baseline-only history blocks this feedback.

The 529 HLoc replacements comprise 189 corrections, seven introduced errors,
and 333 failure-to-failure changes at the primary criterion.
Translation error decreases in 521 of these replacements.

%% file: hpr_sensitivity_table.tex
\begin{table*}[t]
\centering\small
\setlength{\tabcolsep}{2pt}\renewcommand{\arraystretch}{1.0}
\caption{HLoc Sensitivity, Candidate Screening, and Calibration Uncertainty}
\label{tab:calibration-robustness}
\begin{minipage}[t]{0.485\textwidth}
\centering (a) Forecasting: primary $C/H$\par\smallskip
\begin{tabular*}{\linewidth}{@{\extracolsep{\fill}}lccc@{}}
\toprule
Setting & B1 & Night & Hyundai\\
\midrule
Default HPR & $97/6$ & $76/1$ & $16/0$\\
Window: 4 positions & $68/4$ & $62/0$ & $6/0$\\
Window: 7 positions & $103/18$ & $70/7$ & $17/1$\\
Residual factor: 1 & $66/3$ & $69/2$ & $8/0$\\
Residual factor: 3 & $100/6$ & $78/1$ & $16/0$\\
Time/frame indexing & $99/7$ & $76/1$ & $13/0$\\
\bottomrule\end{tabular*}
\par\smallskip{\raggedright\footnotesize One setting changes per row; $\tau$ is recalibrated on B2
under $H\leq5$. Slack 0 or 0.10~m gives default counts.\par}
\par\medskip (b) Screened candidates: $C/H$\par\smallskip
\begin{tabular*}{\linewidth}{@{\extracolsep{\fill}}lccc@{}}
\toprule
Selector, criterion & B1 & Night & Hyundai\\
\midrule
HPR, primary & $95/6$ & $66/1$ & $16/0$\\
HPR, strict & $65/5$ & $15/0$ & $13/0$\\
LTS-only, primary & $68/3$ & $59/2$ & $7/0$\\
LTS-only, strict & $48/3$ & $13/0$ & $5/0$\\
\bottomrule\end{tabular*}
\par\smallskip{\raggedright\footnotesize Screening radius: 1,000~m.
Histories and B2 thresholds are recomputed;
$C/H$ is relative to screened maximum-inlier outputs.\par}
\end{minipage}\hfill
\begin{minipage}[t]{0.485\textwidth}
\centering (c) Cross-calibration: HPR minus LTS-only\par\smallskip
\begin{tabular*}{\linewidth}{@{\extracolsep{\fill}}lcccc@{}}
\toprule Source ($\tau_{\rm HPR}$, $\tau_{\rm LTS}$) & B2 & B1 & Night & Hyundai\\\midrule
B2 (0.8991, 0.9575) & -- & $23\ (14)$ & $7\ (2)$ & $8\ (8)$\\
B1 (0.8235, 0.8516) & $11\ (8)$ & -- & $1\ (0)$ & $1\ (0)$\\
Night (0.4998, 0.5361) & $29\ (25)$ & $36\ (30)$ & -- & $2\ (1)$\\
Hyundai (0.4524, 0.6243) & $9\ (7)$ & $10\ (7)$ & $-9\ (0)$ & --\\
\bottomrule\end{tabular*}
\par\smallskip{\raggedright\footnotesize Primary (strict) net-success differences;
each method uses its source-calibrated threshold. Limits:
B2 5, B1 14, night 38, Hyundai 9; --~marks calibration.\par}
\par\medskip (d) HPR minus LTS-only: primary 95\% ranges\par\smallskip
\begin{tabular*}{\linewidth}{@{\extracolsep{\fill}}lccc@{}}
\toprule
Thresholds and measure & B1 & Night & Hyundai\\
\midrule
Fixed $\tau$: recall (pp) & $[0.46, 1.37]$ & $[0.03, 0.19]$ & $[0.11, 0.91]$\\
Recalibrated $\tau$: net count & $[-7, 67]$ & $[-2, 42]$ & $[-1, 13]$\\
Recalibrated $\tau$: error count & $[-2, 13]$ & $[-1, 4]$ & $[0, 0]$\\
\bottomrule\end{tabular*}
\par\smallskip{\raggedright\footnotesize Fixed thresholds use each method's B2 value:
0.8991 for HPR and 0.9575 for LTS-only; recall intervals are in
percentage points (pp). Recalibrated: central ranges after B2 resampling, targets fixed.
Mean block lengths: $\lceil\sqrt{M_s}\rceil$ (Section~\ref{sec:calibration-robustness}).\par}
\end{minipage}
\end{table*}

%% file: hpr_confidence_transfer_table.tex
\begin{table}[t]
\centering\small\setlength{\tabcolsep}{1.1pt}\renewcommand{\arraystretch}{1.04}
\caption{Confidence Selection and History-Based Recovery}
\label{tab:confidence-transfer}
\textnormal{(a) Full-B2 fitting and calibration}\par\smallskip
\begin{tabular*}{\columnwidth}{@{\extracolsep{\fill}}l*{6}{c}@{}}
\toprule
& \multicolumn{2}{c}{Confidence (cal.)} & \multicolumn{4}{c}{Confidence + HPR}\\
& & & \multicolumn{2}{c}{Unchanged $\tau$} & \multicolumn{2}{c}{Recalibrated $\tau$}\\
Configuration & P & S & P & S & P & S\\\midrule
HLoc B1 & 115/3 & 90/1 & 164/3 & 116/1 & 138/3 & 103/1\\
HLoc night & 48/15 & 22/7 & 89/4 & 32/6 & 69/7 & 28/6\\
HLoc Hyundai & 14/19 & 11/20 & 16/7 & 13/8 & 14/17 & 11/18\\
\midrule
MASt3R & 21/1,532 & 0/190 & 21/1,522 & 0/189 & 21/1,532 & 0/190\\
DUNE & 15/1,372 & 0/235 & 16/1,354 & 0/233 & 15/1,371 & 0/235\\
\midrule
JIST raw & 15/0 & 14/0 & 23/2 & 19/2 & 18/1 & 16/1\\
JIST $T{=}1$ & 12/0 & 12/0 & 17/1 & 17/1 & 16/1 & 16/1\\
JIST $T{\leq}5$ & 18/0 & 15/0 & 24/3 & 19/3 & 22/0 & 18/0\\
\bottomrule\end{tabular*}
\par\smallskip\parbox{\columnwidth}{\footnotesize Primary (P) and strict (S) $C/H$.
HPR uses $\tau=0.8991$ or recalibrates to $0.9760$.
MASt3R/DUNE: Hyundai, HLoc-B2 confidence without refitting.}
\par\medskip\textnormal{(b) Separate B2 captures: primary $C/H$}\par\smallskip
\begin{tabular*}{\columnwidth}{@{\extracolsep{\fill}}l*{6}{c}@{}}
\toprule
& \multicolumn{2}{c}{B1} & \multicolumn{2}{c}{Night} & \multicolumn{2}{c}{Hyundai}\\
Fit$\to$cal. (limit) & Conf. & Comb. & Conf. & Comb. & Conf. & Comb.\\\midrule
$1\to2$ (5) & 121/4 & 173/15 & 57/7 & 98/4 & 14/0 & 18/0\\
$1\to2$ (2) & 106/1 & 160/3 & 37/3 & 87/2 & 14/0 & 16/0\\
$2\to1$ (5) & 119/16 & 172/9 & 52/64 & 95/19 & 13/83 & 18/37\\
$2\to1$ (2) & 80/8 & 103/7 & 19/40 & 52/26 & 12/65 & 12/61\\
\bottomrule\end{tabular*}
\par\smallskip\parbox{\columnwidth}{\footnotesize Conf.: confidence; Comb.: confidence followed by recalibrated HPR.
All counts are relative to maximum inliers; in (a), recalibrated $\tau$ has a five-error B2 limit.
$1\to2$: fit on capture 1, calibrate on 2.
Limits constrain total calibration errors;
2 is the rate-matched limit.}
\end{table}

%% file: hpr_lamar_table.tex
\begin{table}[t]
\centering\small\setlength{\tabcolsep}{2pt}
\caption{Fixed-Threshold Transfer to LaMAR: Camera-Image $C/H$}
\label{tab:lamar}
\begin{tabular*}{\columnwidth}{@{\extracolsep{\fill}}lccccc@{}}
\toprule
Scene, device & $A/F$ & Primary & Strict & \multicolumn{2}{c}{LaMAR criteria}\\
& & 5 m/$10^\circ$ & 0.25 m/$2^\circ$ & 1 m/$5^\circ$ & 0.1 m/$1^\circ$\\\midrule
CAB phone & 13/77 & 4/0 & 2/0 & 2/0 & 1/0\\
CAB HL & 84/108 & 9/0 & 5/0 & 9/0 & 2/0\\
HGE phone & 10/33 & 1/0 & 1/0 & 1/0 & 1/0\\
HGE HL & 63/192 & 2/12 & 0/8 & 2/12 & 0/4\\
LIN phone & 2/18 & 0/0 & 0/0 & 0/0 & 0/0\\
LIN HL & 0/4 & 0/0 & 0/0 & 0/0 & 0/0\\
\midrule
Pooled & 172/432 & 16/12 & 8/8 & 14/12 & 4/4\\
\bottomrule\end{tabular*}
\par\smallskip\parbox{\columnwidth}{\footnotesize HL: HoloLens. $A/F$: recoverable/all primary baseline failures.
The 12 primary HGE HL errors arise from three rig decisions.}
\end{table}

%% file: hpr_operating_table.tex
\begin{table}[t]
\centering\small\setlength{\tabcolsep}{2pt}
\caption{HLoc Recovery Under Sparse Updates and Past-GT History}
\label{tab:operating}
\begin{tabular*}{\columnwidth}{@{\extracolsep{\fill}}l*{6}{c}@{}}
\toprule
& \multicolumn{2}{c}{B1} & \multicolumn{2}{c}{Night} & \multicolumn{2}{c}{Hyundai}\\
Stride & P & S & P & S & P & S\\\midrule
1 & 97/6 & 66/5 & 76/1 & 18/0 & 16/0 & 14/0\\
2 & 63/6 & 42/6 & 65/0 & 18/0 & 13/0 & 11/0\\
4 & 25/10 & 14/8 & 48/1 & 11/0 & 3/0 & 2/0\\
8 & 15/11 & 10/9 & 40/0 & 11/0 & 5/0 & 4/0\\
\midrule
Past GT & 167/0 & 105/0 & 90/0 & 22/0 & 16/0 & 14/0\\
\bottomrule\end{tabular*}
\par\smallskip\parbox{\columnwidth}{\footnotesize Cells: $C/H$ at primary (P) and strict (S) criteria.
Fixed B2 threshold. Strides pool all starting offsets with separate histories,
preserving each target's query count. Past GT uses stride 1 and substitutes only
past positions, preserving baseline update availability.}
\end{table}

%% file: sections/final_conclusion.tex
\section{Conclusion}\label{sec:conclusion}
Coarse forecasts can select accurate current-query poses when candidates are well
separated and recent localization history remains informative.
HPR needs up to five past baseline centers, no odometry or learned model, and
below one millisecond (median) per query on HLoc.
It improves 5~m/$10^\circ$ recall in all 12 main configurations with one
development-set threshold and few introduced errors. Recalibrating HPR after confidence-based selection
adds net recovery on HLoc.
The multi-forecast test permits lower thresholds under matched calibration error budgets;
HPR exceeds best-fit-only selection
in net primary recovery in 11 of 12 HLoc
cross-calibration directions.
Gains weaken with less frequent localization and consecutive history errors,
can reverse when localization resumes after long output gaps, and vary on LaMAR. Ground-truth histories remove all HLoc introduced errors,
suggesting that odometry-aided histories could reduce them.